\documentclass[conference,10pt,letterpaper]{IEEEtran}
\usepackage[T1]{fontenc}

\ifCLASSINFOpdf
\else
\fi
\usepackage{amsmath}
\usepackage{amssymb,latexsym}
\usepackage{latexsym}
\usepackage{graphicx}
\usepackage{subfigure}
\usepackage[noadjust]{cite}
\usepackage{epsfig}
\usepackage{verbatim}
\usepackage{arcs}
\usepackage{setspace}
\usepackage{bm}
\usepackage{cite}
\usepackage[ruled,vlined,linesnumbered]{algorithm2e}
\usepackage{algpseudocode}
\usepackage{wrapfig}
\usepackage{amsmath}
\usepackage{autobreak}
\usepackage{enumitem}
\usepackage{color} 

\newcommand{\ca}{\mathcal}
\newcommand{\ba}{\begin{aligned}}
\newcommand{\ea}{\end{aligned}}
\newtheorem{thm}{Theorem}
\newtheorem{lm}{Lemma}
\newtheorem{df}{Definition}
\newtheorem{as}{Assumption}

\newtheorem{rmk}{Remark}

\IEEEoverridecommandlockouts

\begin{document}
%
\title{Truthful Incentive Mechanism for Federated Learning with Crowdsourced Data Labeling}

%
%
%
\author{\IEEEauthorblockN{Yuxi Zhao,
Xiaowen Gong, Shiwen Mao
}
\IEEEauthorblockA{Department of Electrical and Computer Engineering\\
Auburn University,
Auburn, AL 36849\\
Email: \{yzz0171,xgong\}@auburn.edu, smao@ieee.org}
\thanks{The work of Y. Zhao and X. Gong was supported by the startup fund of X. Gong from Auburn University, and U.S. NSF grant ECCS-2121215. The work of S. Mao was supported in part by U.S. NSF grant CNS-2107190. }
}

\maketitle

\begin{abstract}
Federated learning (FL) has recently emerged as a promising paradigm that trains machine learning (ML) models on clients' devices in a distributed manner without the need of transmitting clients' data to the FL server. In many applications of ML (e.g., image classification), the labels of training data need to be generated manually by human agents (e.g., recognizing and annotating objects in an image), which are usually costly and error-prone. In this paper, we study FL with crowdsourced data labeling where the local data of each participating client of FL are labeled manually by the client. We consider the strategic behavior of clients who may not make desired effort in their local data labeling and local model computation (quantified by the mini-batch size used in the stochastic gradient computation), and may misreport their local models to the FL server. We first characterize the performance bounds on the training loss as a function of clients' data labeling effort, local computation effort, and reported local models, which reveal the impacts of these factors on the training loss. With these insights, we devise Labeling and Computation Effort and local Model Elicitation (LCEME) mechanisms which incentivize strategic clients to make truthful efforts as desired by the server in local data labeling and local model computation, and also report true local models to the server. The truthful design of the LCEME mechanism exploits the non-trivial dependence of the training loss on clients' hidden efforts and private local models, and overcomes the intricate coupling in the joint elicitation of clients' efforts and local models. Under the LCEME mechanism, we characterize the server's optimal local computation effort assignments and analyze their performance. We evaluate the proposed FL algorithms with crowdsourced data labeling and the LCEME mechanism for the MNIST-based hand-written digit classification. The results corroborate the improved learning accuracy and cost-effectiveness of the proposed approaches.
\end{abstract}


%
\IEEEpeerreviewmaketitle

 \section{Introduction}

Federated learning (FL)~\cite{mcmahan17} is an emerging and promising ML paradigm, which performs the training of ML models in a distributed manner. Instead of transmitting data from a potentially large number of devices to a central server in the edge or cloud for training, FL allows the data to remain at devices (such as smartphone), and trains a global ML model on the server by collecting and aggregating model updates locally computed on each device based on her local data. One significant advantage of using FL is to preserve the privacy of individual device’s data. Moreover, since only local ML model updates, instead of local data, are sent to the server, the communication costs can be greatly reduced. Furthermore, FL can exploit the substantial computation capabilities of ubiquitous smart devices, which are often under-utilized. As a result, FL can achieve collaborative intelligence, which can enable many AI applications based on networked systems, such as connected and autonomous vehicles, collaborative robots, multi-user virtual/mixed reality.

Recent studies on FL typically focus on supervised learning, which requires a large amount of training data with data labels in the learning process. In many applications of ML, data labels have to be generated manually by human users. For example, for image classification, the object in an image should be recognized and annotated by a human user as the label of the image data. Therefore, as FL does not allow a client to share her local data with the server or other clients, to participate in FL, a client needs to manually label her local data (e.g., images), before she can compute local model updates from her locally labeled data. 

However, data labels generated by human clients of FL are subject to errors. For example, a client may misclassify a dog as a cat. As a result, this incorrect data label will lead to error in the local model, and thus error in the global model obtained by the FL server. Moreover, the labeling error rate of a client generally varies for different clients, depending on the client's knowledge level of the labeling task. For example, a client who is familiar with dogs will have a lower labeling error rate than another client who is not. Furthermore, the accuracy of data labels is also affected by a client's effort made in the data labeling task. The data label error rate will be low when the client makes much effort in labeling the data, and otherwise is high when the client makes little or no effort. For example, a client may make no effort in image classification by randomly guessing the object in an image without actually recognizing~it.

While a client's effort impacts the accuracy of her data labels, the effort can be her hidden action that is only known by the client herself and cannot be observed by the FL server. Due to the inaccurate nature of data labels, a strategic client may label her local data arbitrarily without making effort in data labeling, while the server will not be able to verify whether effort is actually made or not. Moreover, the effort made by a client in computing her local model update, which can be quantified by the mini-batch size used by the client in stochastic gradient descent, can also be the client's hidden action that cannot be verified by the server. As a result, {a client may have incentive to compute her local update with a small mini-batch size so as to reduce her resources used in local computation.} Furthermore, the local model computed by a client from her local data can also be her private information that she can manipulate in favor of herself, {e.g., a client may increase or decrease her true local model and report it to the server.}

In the presence of such strategic clients with hidden data labeling and local computation efforts and private local models, our goal is to incentivize the clients to make truthful efforts as desired by the FL server and reveal their true local models. Such a truthful incentive mechanism is desirable as it eliminates the possibility of manipulation, which would encourage clients to participate in FL. More importantly, the truthful elicitation of clients' efforts and local models ensures that the FL server can obtain a global model with high and guaranteed accuracy from the learning process, which is a key performance metric of FL. 

The joint elicitation of data labeling effort, local computation effort, and local models for FL calls for a new design that is very different from existing truthful mechanisms. First, the training loss of the global model obtained from FL has a non-trivial dependence on clients' exerted efforts and reported models. As a result, existing incentive mechanisms for effort and data elicitation do not work for the problem here. Second, due to the complex relationship between the impacts of labeling effort, computation effort, and local models on the training loss, the joint elicitation of effort and models needs to overcome the coupling therein. Third, given the truthful incentive mechanism for effort and model elicitation, the FL server needs to determine how much effort should be made by each client, in order to maximize the server's payoff. 


The main contributions of this paper are as follows.
\vspace{-0.1cm}
\begin{itemize}[leftmargin=*]
  \item We propose an FL framework with crowdsourced data labeling based on a truthful incentive mechanism, where the labels of a client's local training data for FL are manually generated by the human client and are subject to errors. We consider strategic clients whose actual efforts in data labeling and local model computation as well as actual local models cannot be verified by the FL server.

  
  \item We first characterize the performance bounds on the training loss as a function of clients’ data labeling effort, local computation effort (quantified by the mini-batch size), and reported local models. It shows that the labeling and computation efforts as well as the reported models have non-trivial impacts on the training loss. Based on the obtained insights, we develop the Labeling and Computation Effort and Local model Elicitation (LCEME) mechanism which incentivize clients to truthfully make efforts in data labeling and local computation, and report local models. The truthful design of the LCEME mechanism overcomes the intricate coupling in the joint elicitation of labeling effort, computation effort, and local models. Based on the LCEME mechanism, we then characterize the optimal computation effort assignment for maximizing the FL server's payoff.
  
  \item We evaluate the proposed FL with crowdsourced data labeling for the MNIST-based hand-written digit recognition. The results demonstrate that the proposed algorithms outperform the methods that do not consider data labeling errors or do not use an incentive mechanism.
\end{itemize}

\vspace{-0.1cm}
The remainder of this paper is organized as follows. Section \ref{sc:relate} reviews the related work. In Section \ref{sc:model}, we describe the system model and formulate the problem of incentive mechanism design. In Section \ref{sc:basic}, we study the performance bound on the training loss. In Section \ref{sc:design}, we devise the LCEME mechanism and the server's optimal effort allocation. Simulation results are presented in Section~\ref{sc:sim}. Section~\ref{sc:concl} concludes this paper.
\vspace{-0.6cm}

\section{Related Work}\label{sc:relate}

\begin{figure*}
\centering
\includegraphics[width=0.78\textwidth]{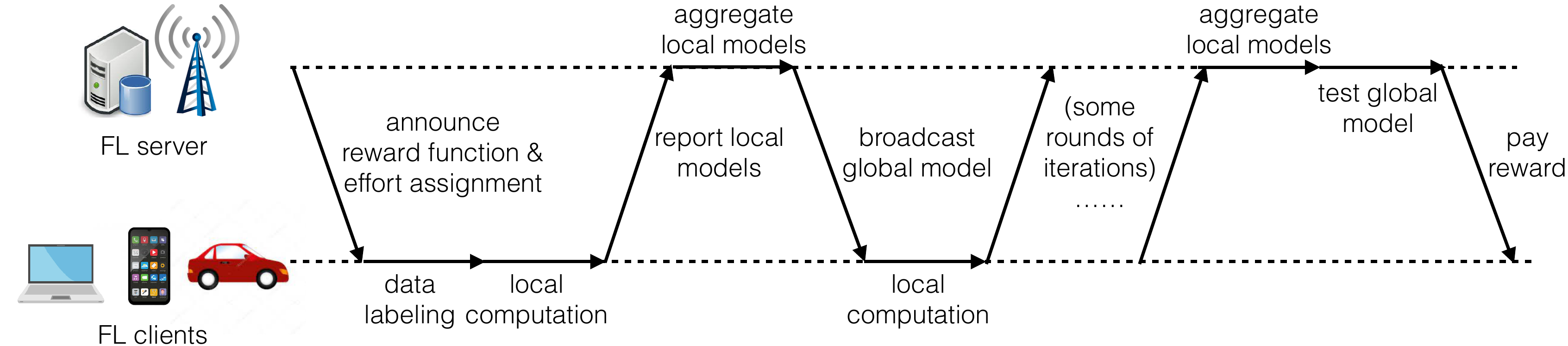}
\caption{Schedule of FL with crowdsourced data labeling based on a truthful incentive mechanism.}
\label{fg:FL-sys}
\vspace{-0.3cm}
\end{figure*}
\noindent\textbf{Incentive Mechanism for Federated Learning.} FL has emerged as a disruptive computing paradigm for ML by democratizing the learning process to potentially many individual devices. Most existing studies on FL have focused on algorithm design for FL, such as for reducing the local model drifts across non-IID clients and participating clients selection. Meanwhile, there have been several recent works on computation and communication resource allocation for FL~\cite{tran19,yang19,chen20,xu20,shi20,ren20,wang2021device,zhang2021federated}. On the other hand, a few recent works studied incentive mechanisms~\cite{sim20icml,yu20is,sun21jsac,zhang21jsac,pandey20twc,jiao20tmc,donahue21aaai,donahue21nips,kang19iot,ding20jsac,zhang2021faithful} for FL that take into account participating clients' strategic behavior. In particular, most of these works considered compensating clients' communication and computation costs with an economic approach, such as Stackelberg game~\cite{pandey20twc}, auction theory~\cite{jiao20tmc}, cooperative game~\cite{donahue21aaai,donahue21nips}, and contract theory~\cite{kang19iot,ding20jsac}. However, all these prior works have focused on either incentivizing clients' participation via cost compensation, or truthfully eliciting clients' participation costs. {\cite{zhang2021faithful} proposed VCG-based mechanisms that incentivize clients to truthfully report their local models.} {In contrast, this paper studies incentive mechanisms for truthful elicitation of clients' local models as well as their efforts in data labeling and local computation.}

\noindent\textbf{Truthful Incentive Mechanism for Effort and Data Elicitation.} There have been lots of research on incentive mechanisms for various applications of data collection and processing, particularly for data crowdsourcing \cite{duan12,yang12,koutsopoulos13,feng14,tarable15,shah15,luo15,wang16,pu16,zhang16}. Many incentive mechanisms incentivize agents to truthfully reveal their participating cost, where the cost is considered to be private for an agent that may not be revealed truthfully without appropriate incentive. There have been studies on truthful mechanism design for hidden efforts in economics literature~\cite{bolton05}, which is concerned with strategic agents that can make hidden efforts not desired by a principal who recruits the agents to work on a task. A few recent works have studied this problem in the context of crowdsourcing~\cite{dasgupta13,cai15,luo15,liu16,liu17}. Mechanism design for truthful elicitation of strategic agents' data (e.g., opinions) has been extensively studied in various applications (e.g., \cite{prelec04}), more recently for crowdsourcing~\cite{dasgupta13,luo15,jin17mobihoc,liu16,liu17}. The data of an agent can be private information that the agent can manipulate in favor of her benefit. Different from existing works, in this paper, {we focus on FL and aim to design truthful mechanisms that jointly elicit clients' hidden efforts in data labeling and local computation and private local models. The truthful mechanism design is non-trivially different from existing works, due to 1) complex dependence of the training loss on clients' data labeling and local computation efforts and local models; 2) intricate coupling in joint elicitation of the clients' efforts and models.}
\vspace{-0.2cm}

\section{System Model and Problem Formulation}\label{sc:model} 
In this section, we first describe a FL system with crowdsourced data labeling based on a truthful incentive mechanism (as illustrated in Fig.~\ref{fg:FL-sys}), and then present the design objectives for truthful incentive mechanisms.
\vspace{-0.1cm}
\subsection{FL with Crowdsourced Data Labeling}
\vspace{-0.1cm}
Consider the following FL problem:
\vspace{-0.2cm}
\begin{equation}
    \min_\mathbf{w} \  F(\mathbf{w})\triangleq \sum_{i=1}^Np_iF_i(\mathbf{w}),
    \label{eq:FL_obj}
    \vspace{-0.2cm}
\end{equation}
where $F_i(\mathbf{w})$ is defined by
\vspace{-0.2cm}
\begin{displaymath}
    F_i(\mathbf{w})\triangleq \frac{1}{\tilde{D}_i}\sum_{m=1}^{\tilde{D}_i}f(\mathbf{w};\xi_m^i),
    \vspace{-0.2cm}
\end{displaymath}
$f(\cdot)$ is the per-sample loss function of client $i$, $N\triangleq|\mathcal{N}|$ is the number of clients, $p_i$ is the weight of client $i$, $\sum_{i\in\mathcal{N}}p_i\triangleq 1$,  $\ca{D}_i=\{\xi_1^i,\xi_2^i,\dots, \xi_{\tilde{D}_i}^i\}$ is client $i$'s local dataset for updating the model parameter, and ${D}\triangleq \sum_{i=1}^N \tilde{D}_i$. Without loss of generality, for ease of exposition, we assume that all clients have the same per-sample loss function $f(\cdot)$.

\noindent\textbf{Data Labeling.} To participate in FL, each client needs to have a labeled local dataset $\mathcal{D}_i$. In this paper, we assume the clients collaboratively train for classification tasks, where each client needs to label her local dataset (i.e., classify the local data samples based on the features of data). After finding the classification labels, each client $i\in\mathcal{N}$ obtains the local dataset $\mathcal{D}_i$.



For simplicity, we assume that each client has two strategies for the labeling effort $e_i\in\{0,1\}$, where $e_i=1$ and $e_i=0$ indicate making and not making effort, respectively. If client $i$ makes effort, then the labels in her dataset are correct; otherwise, the labels are randomly selected from all possible classes without considering the corresponding features. We know that an ML model trained on a correctly labeled dataset is more likely to make useful predictions than a model trained on incorrectly labeled data. Therefore, making effort $e_i=1$ means higher accuracy of the trained model than not making effort (We prove this intuition in Section \ref{sc:basic}.). We assume that every client can fully control the amount of effort they make, and the server does not have such information.

\noindent\textbf{Local Model Computation.} In each round of FL, clients communicate their local updates to the server and receive the updated global model from the server. In round $t$,\footnote{We use $t$ and $h$ as the index of communication rounds and local iterations, respectively. The subscript $(t,h)$ denotes the $h$th local iteration in round $t$.} each client $i$ receives the global model $\mathbf{w}_{t-1}$ from the server, sets $\mathbf{w}^i_{t,0}=\mathbf{w}_{t-1}$, and then performs $H$ local iterations of SGD. In the $h$th local iteration, client $i$ computes the average gradient $g_{t,h-1}^i$ of the loss function using a mini-batch of $D_i$ data samples randomly drawn from her local dataset $\ca{D}_i$. Then client $i$ updates her local model as
 \vspace{-0.2cm}
\begin{displaymath}\label{update}
\mathbf{w}^i_{t,h}=\mathbf{w}^i_{t,h-1}-\eta g^i_{t,h-1},
\vspace{-0.3cm}
\end{displaymath}
where  
 \begin{displaymath}
 g^i_{t,h-1}\triangleq \frac{1}{{D}_i}\sum_{j=1}^{D_i} \nabla f(\mathbf{w},\xi^{i,j}_{t,h}) ,
 \vspace{-0.3cm}
 \end{displaymath}
$\eta$ is the step size, and $\xi^{i,j}_{t,h}$ is the $j$th data sample randomly drawn from client $i$'s local dataset $\ca{D}_i$. After $H$ local iterations, client $i$ sends her local update $\mathbf{w}_{t,H}^i$ for round $t$ to the server. 

The computation effort $D_i$ represents the mini-batch size client $i$ uses to update her local model in each round. Due to the randomness of data sampling for computing the update in SGD, the computed gradient of a client could deviate from the expected gradient, and thus slow down the convergence of the FL global model. It has been proved that the larger the mini-batch size $D_i$, the lower the variance of her local update \cite{dekel12optimal}. Thus, a local update computed with a larger mini-batch size benefits the FL training. 

At the end of round $t$, the server aggregates clients' local models and updates the global model as
\vspace{-0.15cm}
\begin{displaymath}
    \mathbf{w}_t = \sum_{i=1}^Np_i \mathbf{w}_{t,H}^i.
    \vspace{-0.2cm}
\end{displaymath}

\noindent\textbf{Effort Assignment.} Before data labeling and local computation, the server assigns the labeling effort $e_i^\prime$ and computation effort $D_i^\prime$ to each client $i$ and notifies client $i$ of $e_i^\prime$ and $D_i^\prime$. The labeling effort $e_i^\prime\in\{0,1\}$ indicates whether the server desires client $i$ to make effort in labeling, and the computation effort $D_i^\prime$ indicates the mini-batch size that the server desires client $i$ to use to update her local model in each round. Clients' effort assignments generally vary for different clients due to their diverse characteristics (e.g., weight in model aggregation, computation capability). 

After being assigned effort $e_i^\prime$, each client $i$ generates labels for the local dataset by making actual effort $e_i$. Since $e_i$ is a hidden action of client $i$, it is possible that client $i$ manipulates $e_i$ against the assignment $e_i^\prime$ to her own advantage such that $e_i\neq e_i^\prime$. 

Furthermore, a client incurs a computation cost (measured by the computation time, energy consumption, etc.) for computing a local model update, which depends on the computation capability of the client and the mini-batch size used to compute the update. Thus, client $i$ may also have incentive to manipulate $D_i$ against the assignment $D_i^\prime$ to her own advantage such that $D_i\neq D_i^\prime$. 

\noindent\textbf{Local Model Reporting.} When reporting the local model to the server, a client $i$ may have incentive to misreport her local model to her own advantage, i.e., 
\vspace{-0.2cm}
\begin{displaymath}\label{update}
\mathbf{w}^i_{t}=\mathbf{w}_{t-1}-\gamma_i\eta g^i_{t-1},
\vspace{-0.15cm}
\end{displaymath}
where $\gamma_i\geq 0$, $\forall i\in \mathcal{N}$ is the model reporting coefficient, which is the multiple of the gradient  client $i$ uses to update her local model\footnote{{In this paper, we assume that clients' strategies do not change over time in FL training.}}. When $\gamma_i=1$, client $i$ reports the actual local model to the server, which is desired by the FL server. When $\gamma_i\neq 1$, the gradient is reduced or increased. In this case, the trained model of FL will be affected, and thus the training loss $F(\mathbf{w})$. It is possible that client $i$ manipulates $\gamma_i$ to her own advantage such that $\gamma_i \neq 1$. 

\subsection{Truthful Incentive Mechanism for FL}

{After the training process, the FL server tests the trained global model of FL to a data sample $\xi$ randomly drawn from a testing dataset $\mathcal{D}_0$. It is commonly assumed in existing studies that the FL server can test the trained FL model (e.g.,~\cite{kang19iot, song2022personalized}). Then the server can determine each client's reward based on the testing loss $f(\mathbf{w}_T,\xi)$ observed for the testing data sample $\xi$. Note that the server only needs to test the trained model to a single random data sample from $\mathcal{D}_0$. For example, the testing can be performed when the server applies the trained model to an unseen data sample for inference/prediction and observes its true label later.}

{Based on the testing loss $f(\mathbf{w}_T,\xi)$, the server pays a reward $r_i$ to each client $i$, according to a certain reward function:
\vspace{-0.2cm}
\begin{equation}
    r_i(e_i^\prime,\boldsymbol{e}_{-i}^\prime,D_i^\prime,\boldsymbol{D}_{-i}^\prime,\gamma_i^\prime, \boldsymbol{\gamma}_{-i}^\prime, f(\mathbf{w}_T,\xi)),
    \label{eq:r_i_s3}
    \vspace{-0.2cm}
\end{equation} where $\boldsymbol{e}_{-i}^\prime$, $\boldsymbol{D}_{-i}^\prime$, and $\boldsymbol{\gamma}_{-i}^\prime$ are other clients' assigned data labeling and computation effort, and model reporting coefficient, respectively. The reward function is pre-defined by the server and announced to all clients before they participate in FL. We can see that the reward depends on not only the assigned efforts and model reporting coefficient but also the testing loss of the final global model.}

Each client $i$'s payoff is the difference between the reward paid by the server and her cost in data labeling and computing her local model, given by
\vspace{-0.2cm}
\begin{equation*}
\ba
   &\ u_i(e_i,\boldsymbol{e}^\prime,D_i, \boldsymbol{D}^\prime,\gamma_i, \boldsymbol{\gamma}^\prime) \triangleq
    \\[-0.2cm]& r_i(e_i^\prime,\boldsymbol{e}_{-i}^\prime,D_i^\prime,\boldsymbol{D}_{-i}^\prime,\gamma_i^\prime, \boldsymbol{\gamma}_{-i}^\prime, f(\mathbf{w}_T,\xi))-c_l^ie_i-\sum_{t=1}^Tc_p^iD_i,
    \ea
    \vspace{-0.2cm}
\end{equation*}
where $\boldsymbol{e}^\prime$, $\boldsymbol{D}^\prime$, and $\boldsymbol{\gamma}^\prime$ are clients' assigned data labeling effort, computation effort, and model reporting coefficient, respectively. The data labeling cost coefficient $c_l^i$ captures the resources consumed by client $i$ if she makes an effort, i.e., $e_i=1$, in data labeling, and the computation cost coefficient $c_p^i$ is client $i$'s cost of computing her local update using one data sample. If client $i$ makes no effort in data labeling, i.e., $e_i=0$, there incurs no data labeling cost. {Here we assume that clients have the same data labeling cost coefficient (i.e., $c_l=c_l^i$, $\forall i \in \mathcal{N}$), and the labeling and computation cost coefficients are known to the server. This assumption is reasonable when the costs of labeling a client's dataset and computing using a data sample are determined by uniform market prices {(e.g., in Amazon Mechanical Turk, a usual reward for labeling an image is $\$$0.1)}.} A client's computation cost is affected by her computation cost coefficient $c_p^i$ and computation effort $\boldsymbol{D}^\prime$. We can also relax the restriction of the uniform labeling cost coefficient. Since a client $i$ can only affect the training loss through her actual $e_i$, $D_i$, and $\gamma_i$, we omit the loss function $f(\mathbf{w}_T,\xi)$ in the expression of client $i$'s utility $u_i$. The detailed reward function design will be given in Section~\ref{sc:design}.

The server's payoff $u_0$ is the negative training loss minus the total reward paid to the clients, i.e.,
\vspace{-0.2cm}
\begin{equation}
    u_0(\boldsymbol{e}^\prime, \boldsymbol{D}^\prime,\boldsymbol{\gamma}^\prime, f(\mathbf{w}_T,\xi))\triangleq - f(\mathbf{w}_T,\xi)-\sum_{i\in\mathcal{N}}r_i.
\vspace{-0.2cm}
\end{equation}

Since clients may manipulate their actual efforts and report untruthful local models, the global model may be different from that when clients do not behave truthfully, i.e.,
\vspace{-0.2cm}
\begin{displaymath}
    \mathbf{w}_T|_{\boldsymbol{e}^\prime,\boldsymbol{D}^\prime, \boldsymbol{\gamma}^\prime}\neq \mathbf{w}_T|_{\boldsymbol{e},\boldsymbol{D},\boldsymbol{\gamma}}.
    \vspace{-0.2cm}
\end{displaymath}
This means that the final global model obtained with efforts and reported local model manipulation cannot solve the FL problem given in \eqref{eq:FL_obj}. Nevertheless, the training loss of FL is also affected, i.e.,
\vspace{-0.2cm}
\begin{displaymath}
    F(\mathbf{w}_T)|_{\boldsymbol{e}^\prime,\boldsymbol{D}^\prime, \boldsymbol{\gamma}^\prime}\neq F(\mathbf{w}_T)|_{\boldsymbol{e},\boldsymbol{D},\boldsymbol{\gamma}}.
    \vspace{-0.2cm}
\end{displaymath}

Furthermore, some clients’ manipulation would discourage other clients to participate in FL. For the reasons discussed above, here we aim to design a mechanism that can incentivize clients to make data labeling and computation efforts as the server desired and upload their actual local models. This can be achieved by properly defining the reward function $r_i$. The truthful mechanism should have the following features:
\begin{df}\label{df:nq}
	A mechanism achieves truthful strategies of all clients as a \textit{Nash equilibrium} (NE) if, given that all other clients truthfully make data labeling and computation effort as the server desired and upload their actual local models, the best strategy for client $i$ to maximize her payoff is to behave truthfully, i.e.,
    	\begin{align}
    	    E[u_i(e_i^\prime,\boldsymbol{e}_{-i}^\prime,& D_i^\prime,\boldsymbol{D}_{-i}^\prime, \gamma_i^\prime,\boldsymbol{\gamma}_{-i}^\prime)]\geq \nonumber\\& \hspace{-0.5cm}E[u_i(e_i,\boldsymbol{e}_{-i}^\prime,D_i,\boldsymbol{D}_{-i}^\prime,\gamma_i, \boldsymbol{\gamma}_{-i}^\prime)], \forall e_i,D_i,\gamma_i.	\label{eq:truth}
    	\end{align}

\end{df}	

	We should also notice that the payoff of each client $i$ should be non-negative so that the client will have the incentive to participate. This property is known as individual rationality.
	\begin{df}\label{df:ir}
		A mechanism is \textit{individually rational} (IR) if for each client $i$, its expected payoff is non-negative if she behaves truthfully, i.e.,
		\vspace{-0.2cm}	
		\begin{equation}
	 E[u_i(e_i^\prime,\boldsymbol{e}_{-i}^\prime, D_i^\prime,\boldsymbol{D}_{-i}^\prime, \gamma_i^\prime,\boldsymbol{\gamma}_{-i}^\prime)]\geq 0,\forall e_i,D_i,\gamma_i.
		\end{equation}
	\end{df}
	\vspace{-0.2cm}	
%

\section{Training Loss Analysis} \label{sc:basic}
In this section, we characterize the performance bounds on the training loss as a function of clients' data labeling effort, local computation effort, and reported local models, which reveal the impacts of these factors on the training loss.

We first make the following general assumptions on the loss functions $F_1, \dots , F_N$, $\forall i\in \mathcal{N}$.
\begin{as}
$F_1, \dots , F_N$ are $L$-smooth.
\end{as}
\begin{as}
$F_1, \dots , F_N$ are $\mu$-strongly convex.
\end{as}
\begin{as}
The variance of the stochastic gradient of a data sample in a device is bounded: ${E}\left\|\nabla f\left(\mathbf{w}_{t},\xi_m^i\right)-\nabla F_{i}\left(\mathbf{w}_{t}\right)\right\|^{2} \leq \sigma_i ^{2}$, $\forall i\in \mathcal{N}$, $\forall t$.
\end{as}
\begin{as}
The variance of the stochastic gradient of a data sample when the client makes no effort on labeling is bounded: ${E}\left\|\nabla f\left(\mathbf{w}_{t},\xi_m^i\right)\!-\!\nabla f\!\left(\mathbf{w}_{t},{\xi_m^i}^\prime\right)\right\|^{2} \!\!\!\leq \!\beta$, $\forall i\!\in\!\mathcal{N}$, $\forall t$.
\label{as:4}
\end{as}
\begin{as}
The expected squared norm of stochastic gradients is bounded: ${E}\left\|\nabla F_{i} \left(\mathbf{w}_{t}\right)\right\|^{2} \leq G^{2}$, $\forall i\in \mathcal{N}$, $\forall t$.
\end{as}

In Assumption 4, we assume that the variance of the stochastic gradient of a data sample when the client makes no labeling effort is upper bounded, and the bound $\beta$ is known by the server. The server can calculate the bound using the loss function and the range of data's value. Next, we use a simple example to demonstrate how to obtain the bound $\beta$. We use a simple linear regression model to illustrate the convergence problem. Assume that the loss function is given by
\vspace{-0.2cm}	
\begin{displaymath}
f\left(\mathbf{w},\xi_m^i\right)=\frac{1}{2}\|\boldsymbol{x}_m^i\boldsymbol{w}-y_m^i\|^2, \ \ \ \ \forall i\in \mathcal{N}.
\vspace{-0.2cm}	
\end{displaymath}
A data sample with correct and incorrect labels are denoted as $\xi_m^i=(\boldsymbol{x}_m^i,y_m^i)$ and ${\xi_m^i}^\prime=(\boldsymbol{x}_m^i,{y_m^i}^\prime)$, respectively.

The variance of the stochastic gradient of a data sample is 
\vspace{-0.2cm}	
\begin{displaymath}
\ba
&{E}\left\|\nabla f\left(\mathbf{w},\xi_m^i\right)-\nabla f\left(\mathbf{w},{\xi_m^i}^\prime\right)\right\|^{2}
\\=&\|(\boldsymbol{x}_m^i\boldsymbol{w}-y_m^i)\boldsymbol{x}_m^i-(\boldsymbol{x}_m^i\boldsymbol{w}-{y_m^i}^\prime)\boldsymbol{x}_m^i\|^2
\\=&\|({y_m^i}^\prime-y_m^i)\boldsymbol{x}_m^i\|^2\leq 2YX,
\ea
\vspace{-0.2cm}
\end{displaymath}
where ${y_m^i}^2\leq Y$ and $\|\boldsymbol{x}_m^i\|^2\leq X$.
Then we have $\beta=2YX$.

\begin{thm}
Suppose Assumptions 1 to 5 hold, and the step size $\eta \leq \frac{1}{2L}$. Then the FL training loss is bounded above by:
    \begin{align}
    &E[F(\mathbf{w}_{T})-F(\mathbf{w}^*)]  
       \leq{L}(1-\mu \eta)^{TH} E\left\|\mathbf{w}_{0}-\mathbf{w}^{*}\right\|^{2}\nonumber
     \\&  +2{L\eta^2}\sum_{t=1}^{T}\sum_{h=1}^H(1-\mu \eta)^{TH-(t-1)H-h}\nonumber
     \\&\sum_{i\in {\mathcal{N}}} \left({p_i}^2\frac{\sigma_i^2}{D_i}+6L p_id_i+p_i{(1-e_i)}\beta\right.\nonumber
     \\& \left.+ 2p_i\left((\gamma_i\!-\!1)^2\!+\!(H\!-\!1)^2\right)\left(G^2+\frac{\sigma_i^2}{D_i}+(1-e_i)\beta\!\right)\!\right), \label{eq:loss}
    \end{align}
    where $d_i \triangleq E[F_i(\mathbf{w}^*)]-E[F_i(\mathbf{w}_i^*)]$ quantifies the heterogeneity degree of the data held by client $i$ \cite{li20iclr}.
\label{thm:loss}
\end{thm}

The proof is given in Appendix A.

\begin{rmk}
The first term of the training loss bound decreases geometrically with the total number of local iterations $TH$, and is due to that SGD in expectation makes progress towards the optimal solution. The bound is also affected by other factors, i.e., the randomness of data sampling in SGD for computing local updates $p_i^2\frac{\sigma_i^2}{D_i}$, the data heterogeneity of clients' data $p_id_i$, the data labeling effort level of each client $p_i{(1-e_i)}\beta$, the local model misreporting $\gamma_i$, and the number of local iterations per round $H$. We can see that any $\gamma_i\neq 1$, i.e., any client untruthfully reports her local model, increases the training loss bound. Thus, it is desired that all clients report their actual local model (i.e., $\gamma_i=1$, $\forall i\in \mathcal{N}$) to minimize the training loss. Moreover, as the coefficients in the training loss bound depend on the aggregation weight $p_i$, a client with a higher weight $p_i$ has a larger impact on the training loss than that with a lower weight $p_i$. 
\label{rem.datasize}
\end{rmk}

\begin{rmk}
The randomness of data sampling in SGD for computing local updates affects the training loss, which depends on each client's mini-batch size ${D}_i$ in each iteration (i.e., computation effort). We can observe that a larger mini-batch size $D_i$ reduces the training loss. The terms involving $e_i$ depend on the data labeling effort of each client. If client $i$ makes effort in data labeling, these terms equal 0; otherwise, if client $i$ makes no effort in data labeling, these terms equal $p_i\beta$. Thus, it is desirable that all clients make data labeling effort (i.e., $e_i=1$, $\forall i\in \mathcal{N}$) to minimize the training loss.
\end{rmk}

\section{Truthful Incentive Mechanisms for Data Labeling Effort, Local Computation Effort, and Local Model Elicitation}\label{sc:design}
In this section, we propose the LCEME mechanism that satisfies the truthful and IR properties to incentivize clients to make efforts as the server desired and report actual local models. Then, we find the optimal computation effort assignment under the LCEME mechanism that maximizes the server's payoff.
\vspace{-0.2cm}
\subsection{LCEME Mechanism Design}
\vspace{-0.1cm}

We first present the design of the LCEME mechanism.




\begin{df}
Given the data labeling effort assignment $e_i^\prime=1$, the model reporting coefficient assignment $\gamma_i^\prime=1$, and any computation effort assignment $D_i^\prime$, the LCEME mechanism's reward function for client $i$, $\forall i \in \mathcal{N}$, is given by
\vspace{-0.2cm}
\begin{equation}
    \ba
     r_i(e_i^\prime,&\boldsymbol{e}_{-i}^\prime,D_i^\prime,\boldsymbol{D}_{-i}^\prime,\gamma_i^\prime, \boldsymbol{\gamma}_{-i}^\prime, f(\mathbf{w}_T,\xi))
    \\& =\Omega(\boldsymbol{D}^\prime)-\Phi(D_i^\prime) f(\mathbf{w}_T,\xi)+c_l,
    \ea
    \vspace{-0.2cm}
\end{equation}
	where 
	\vspace{-0.3cm}
\begin{displaymath}
\ba
    &\Omega(\boldsymbol{D}^\prime)
    ={\Phi(D_i^\prime)}\left(L(1-\mu \eta)^{TH} E\left\|\mathbf{w}_{0}-\mathbf{w}^{*}\right\|^{2}\right.+
    \\& \left.A\sum_{i\in {\mathcal{N}}}(6L p_id_i+{p_i}^2\frac{\sigma_i^2}{D_i^\prime}+2p_i(H-1)^2(G^2+\frac{\sigma_i^2}{D_i^\prime}))\right)+Tc_p^iD_i^\prime
    ,
\ea
\vspace{-0.2cm}
\end{displaymath}
$\boldsymbol{e}^\prime=\mathbf{1}^{1\times N}$, $\boldsymbol{\gamma}^\prime=\mathbf{1}^{1\times N}$, $\Phi(D_i^\prime)=\frac{{D_i^\prime}^2c_p^iT}{A\sigma_i^2p_i(p_i+2(H-1)^2)}$, $A=2L\eta\frac{1-(1-\mu\eta)^{TH}}{\mu}$, and the assigned computation effort $ D_i^\prime$ satisfies $D_i^\prime\geq \sigma_i\sqrt{\frac{c_lp_i(p_i+2(H-1)^2)}{\beta c_p^iT(1+2(H-1)^2)}}.$
\end{df}

{Note that the reward function depends on the testing loss which is observed by the server. In this paper, for ease of exposition, we assume that the expected testing loss is equal to the training loss. This assumption is reasonable: in practice, the entire training dataset of FL (i.e., $\cup^{N}_{i=1}\mathcal{D}_i$) is often a good representation of the testing dataset $\mathcal{D}_0$, so that the expected testing loss is well approximated by the training loss. Based on this assumption, the expected payoff of client $i$ is given by: 
\vspace{-0.2cm}
\begin{align}
&E_{\xi}[u_i(e_i,\boldsymbol{e}_{-i}^\prime,D_i,\boldsymbol{D}_{-i}^\prime,\gamma_i, \boldsymbol{\gamma}_{-i}^\prime)]\nonumber
\\&= E_{\xi}[r_i(e_i^\prime,\boldsymbol{e}_{-i}^\prime,D_i^\prime,\boldsymbol{D}_{-i}^\prime,\gamma_i^\prime, \boldsymbol{\gamma}_{-i}^\prime,f(\mathbf{w}_T,\xi))]-c_le_i-Tc_p^iD_i\nonumber
\\ &= \Omega(\boldsymbol{D}^\prime)-\Phi(D_i^\prime)F(\mathbf{w}_T)+c_l-c_le_i-Tc_p^iD_i

\label{eq:u_idef}
\end{align}
where $\xi$ is a random data sample drawn from the testing dataset $\mathcal{D}_0$.}

{Next, based on Theorem~\ref{thm:loss}, we approximate the expected training loss $F(\mathbf{w}_T)$ in terms of the optimal training loss $F(\mathbf{w}^*)$ plus the upper bound on the training loss gap given in the right-hand-side of~\eqref{eq:loss}.
Then we assume that each client uses $\hat{u}_i$ as her expected payoff function, where $\hat{u}_i$ is defined as \eqref{eq:u_idef} with $F(\mathbf{w}_T)$ replaced by the right-hand-side of~\eqref{eq:loss} (the optimal training loss term $F(\mathbf{w}^*)$ is omitted as it does not affect the truthful mechanism design). This is a reasonable assumption since 1) a client cannot find the expected training loss $F(\mathbf{w}_T)$, but can find the upper bound in~\eqref{eq:loss}; 2) using the upper bound on the training loss gap can capture the worst case of the client's expected payoff. Therefore, in the rest of this paper, each client determines her strategic behavior for maximizing the payoff function $\hat{u}_i$.}

{Next, we use two theorems to prove that the LCEME mechanism satisfies the truthful and IR properties, with respect to the clients' payoff functions $\hat{u}_i$.}

\begin{thm}\label{thm:NE}
The LCEME mechanism is truthful.
\end{thm}

We show how the LCEME mechanism achieves the truthful property using three lemmas. 
\begin{lm}
Under the LCEME mechanism, given that client $i$ makes any data labeling effort $e_i$ and computation effort $D_i$, her optimal reported local model is her true local model, i.e., $\gamma_i=1$.
\label{lm:model_truthful}
\end{lm}

It can be shown that the expected payoff of client $i$ is a convex function of $\gamma_i$. We can obtain the result of Lemma \ref{lm:model_truthful} by calculating the partial derivative of the expected payoff of client $i$ with respect to $\gamma_i$ and letting the derivative equal 0.

Using Lemma \ref{lm:model_truthful}, we can express client $i$'s approximated expected payoff $\hat{u}_i$ as 
\begin{displaymath}
\ba
& \hat{u}_i(e_i,D_i,D_i^\prime)
=  \Phi(D_i^\prime)A({p_i}^2\frac{\sigma_i^2}{D_i^\prime}+2p_i(H-1)^2\frac{\sigma^2_i}{D_i^\prime})+Tc_p^iD_i^\prime
\\&-\Phi(D_i^\prime)A\left(p_i^2\frac{\sigma^2_i}{D_i}+p_i(1-e_i)\beta\right.
\\&\left.+2p_i(H-1)^2(\frac{\sigma^2_i}{D_i}+(1-e_i)\beta)\right)
+c_l-c_le_i-Tc_p^iD_i
.
\ea
\end{displaymath}

\begin{lm}
  Under the LCEME mechanism, given that clients report their optimal local models $\gamma_i=1$, $\forall i\in \mathcal{N}$, and client $i$ makes any computation effort, client $i$'s optimal actual effort is the desired effort, i.e., $e_i=1$.
  \label{lm:e_truthful}
\end{lm}
 
Then, we show that, when client $i$ makes any labeling effort, her expected payoff is always lower than that when she makes effort:
\vspace{-0.2cm}
\begin{displaymath}
\ba
&\hat{u}_i(1,D_i,D_i^\prime)-\hat{u}_i(e_i,D_i,D_i^\prime)
\\= & \frac{{D_i^\prime}^2c_p^iT(1+2(H-1)^2)}{\sigma_i^2p_i^2(p_i+2(H-1)^2)}p_i(1-e_i)\beta-c_l+c_le_i
\\= &(\frac{{D_i^\prime}^2c_p^iT(1+2(H-1)^2)\beta}{\sigma_i^2p_i(p_i+2(H-1)^2)}-c)(1-e_i)
\\\geq &(c-c)(1-e_i)\geq 0,
\ea
\vspace{-0.2cm}
\end{displaymath}
where the inequality follows from the constraint on $D_i^\prime$.

Using Lemma \ref{lm:model_truthful} and Lemma \ref{lm:e_truthful}, we can express client $i$'s approximated expected payoff $\hat{u}_i$ as
\vspace{-0.2cm}
\begin{displaymath}
\ba
 \hat{u}_i(D_i,D_i^\prime)
 =& -\Phi(D_i^\prime)A(p_i^2\frac{\sigma^2_i}{D_i}+2p_i(H-1)^2\frac{\sigma^2_i}{D_i})-Tc_p^iD_i
\\&+\Phi(D_i^\prime)A({p_i}^2\frac{\sigma_i^2}{D_i^\prime}+2p_i(H-1)^2\frac{\sigma^2_i}{D_i^\prime})+Tc_p^iD_i^\prime
.
\ea
\vspace{-0.2cm}
\end{displaymath}

\begin{lm}
  Given that clients report their optimal local models $\gamma_i=1$ and make effort in data labeling $e_i=1$, $\forall i\in \mathcal{N}$, client $i$'s optimal actual computation effort is the desired computation effort, i.e., $ D_i=D_i^\prime$.
\end{lm}

Now that the expected payoff is a convex function of client $i$'s actual computation effort $D_i$, we can obtain client $i$'s optimal actual computation effort $D_i$ by calculating the partial derivative of the expected payoff of client $i$ with respect to $D_i$ and letting the derivative equal to 0, which is the desired computation effort $D_i^\prime$.

Given the definition of truthful mechanisms (Definition \ref{df:nq}), the LCEME mechanism is truthful.
$\hfill\square$
\begin{thm}\label{thm:IR}
The LCEME mechanism is IR.
\end{thm}

The proof is given in Appendix B.

\begin{rmk}
Here we discuss the rationale of the LCEME mechanism. The server's goal is to incentivize clients to make actual data labeling and computation effort as desired by the server and report their true local models. Thus, client $i$'s reward function $r_i$ should be a function of her actual efforts ($e_i$ and $D_i$) and model report coefficient ($\gamma_i$). Otherwise, clients can deceive the server to gain more rewards. Thus, we design the reward function as a function of the training loss, which has been proved to be determined by clients' actual efforts and model reporting strategies in Theorem \ref{thm:loss}. In the refined reward function, client $i$'s optimal strategy to maximize her expected payoff is to make data labeling and computation efforts as desired by the server and report her actual local model.
\end{rmk}
\vspace{-0.2cm}
\subsection{Optimal Computation Effort Assignment}
\vspace{-0.1cm}
A desirable objective for the server is to find the optimal assignment that maximizes her expected payoff.

\begin{df}
The server's optimal assignment $\boldsymbol{D}^*$ for LCEME mechanism is the assignment function $\boldsymbol{D}^\prime$ that maximizes the server's payoff, i.e.,
\vspace{-0.2cm}
\begin{equation}
\ba
        &\boldsymbol{D}^*\triangleq \arg \max_{\boldsymbol{D}^\prime} E[u_0(\boldsymbol{D}^\prime, f(\mathbf{w}_T,\xi))] 
        \\ &s.t.\ \  D_i^*\geq \sqrt{\frac{c_l\sigma_i^2p_i(p_i+2(H-1)^2)}{\beta c_p^iT(1+2(H-1)^2)}}, \forall i\in \mathcal{N}.
        \label{eq:SO}
\ea
\vspace{-0.1cm}
\end{equation}
\end{df}

The constraint in \eqref{eq:SO} is to make sure that the LCEME mechanism is truthful.

The problem given in \eqref{eq:SO} is equivalent to the problem:
\vspace{-0.2cm}
\begin{equation}
\ba
    \boldsymbol{D}^*\triangleq &\arg \min_{\boldsymbol{D}^\prime} \left\{F(\mathbf{w}_{T})-F(\mathbf{w}^*)+\sum_{i\in \mathcal{N} } r_i\right\},
    \\ s.t. &\ \  D_i^*\geq \sqrt{\frac{c_l\sigma_i^2p_i(p_i+2(H-1)^2)}{\beta c_p^iT(1+2(H-1)^2)}}, \forall i\in \mathcal{N},
    \ea
    \vspace{-0.2cm}
\end{equation}
where $F(\mathbf{w}^*)$ can be seen as a constant.

From the above problem formulation, we observe that there exists a tradeoff between the FL training loss and the server's payment to clients. We know that the training loss reduces when clients use larger mini-batch sizes to compute their local updates from Theorem \ref{thm:loss}. However, using larger mini-batch sizes increases the server's payment. Therefore, we aim to find the optimal computation effort (in the form of mini-batch size) assignment for each client to maximize the server's payoff. 

\begin{thm}
The server's optimal computation effort allocation is given by
\vspace{-0.2cm}
\begin{displaymath}
\ba
    D_i^*=\max &\left\{\sqrt{\frac{A(p_i^2\sigma_i^2+2p_i(H-1)^2)}{ c_p^iT}}, \right.
    \\&\left.\sqrt{\frac{c_l\sigma_i^2p_i(p_i+2(H-1)^2)}{\beta c_p^iT(1+2(H-1)^2)}}\right\}, \forall i\in \mathcal{N}.
\ea
\vspace{-0.3cm}
\end{displaymath} 
\label{thm:SO}
\end{thm}

The proof is given in Appendix C.






\begin{rmk}
From Theorem \ref{thm:SO}, we can see that the server's optimal computation effort for a client $i$ increases with her weight $p_i$ and gradient variance $\sigma_i^2$. This is because when client $i$ has a larger $p_i$ and/or $\sigma_i^2$, the effect of the randomness of her SGD computation per data sample on the global model will be larger. From Theorem \ref{thm:loss}, we know that a larger mini-batch size $D_i$ reduces the randomness of data sampling in SGD. Thus, assigning a larger computation effort for client $i$ can reduce the training loss. We also see that $D_i^*$ decreases as client $i$'s computation cost $c_p^i$ increases. This is because a larger computation cost increases the reward paid by the server. When a client's computation cost is large, the server prefers to allocate a smaller mini-batch size to the client to reduce the payment. We can also show that a client's optimal mini-batch size increases as the number of local iterations $H$ increases. This is because a local update's quality can be improved by using a larger mini-batch size, and thus reduce the error caused by performing multiple local iterations.
\end{rmk}
\section{Simulation Results}\label{sc:sim}
\begin{figure*}[t]
\centering
\begin{minipage}[t]{.32\textwidth}
\centering
\includegraphics[width=1\textwidth]{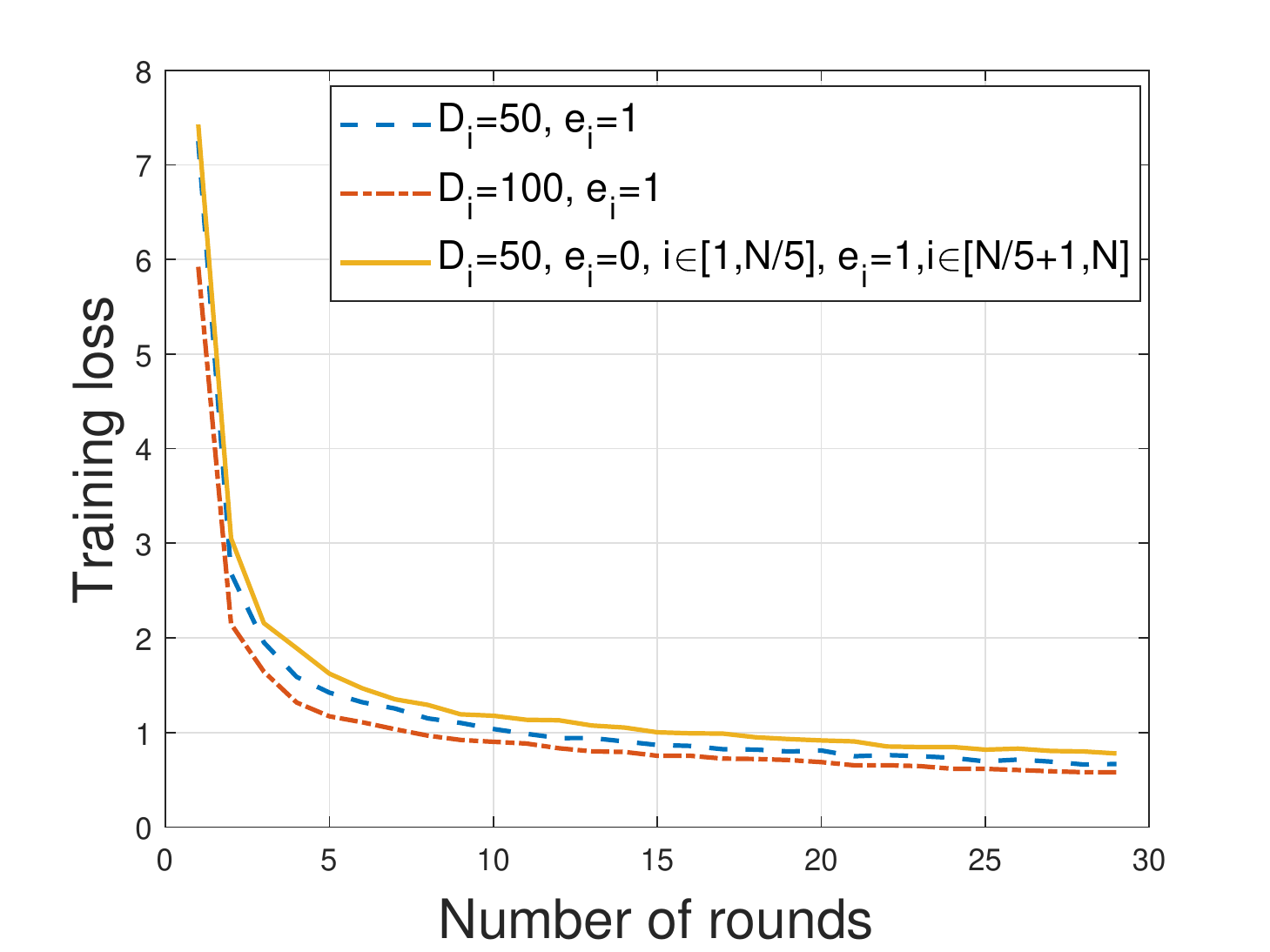}
\caption{Impact of effort level on the training loss.}
\label{fig:ed_loss}
\end{minipage}
\hfill
\begin{minipage}[t]{.32\textwidth}
\centering
\includegraphics[width=1\textwidth]{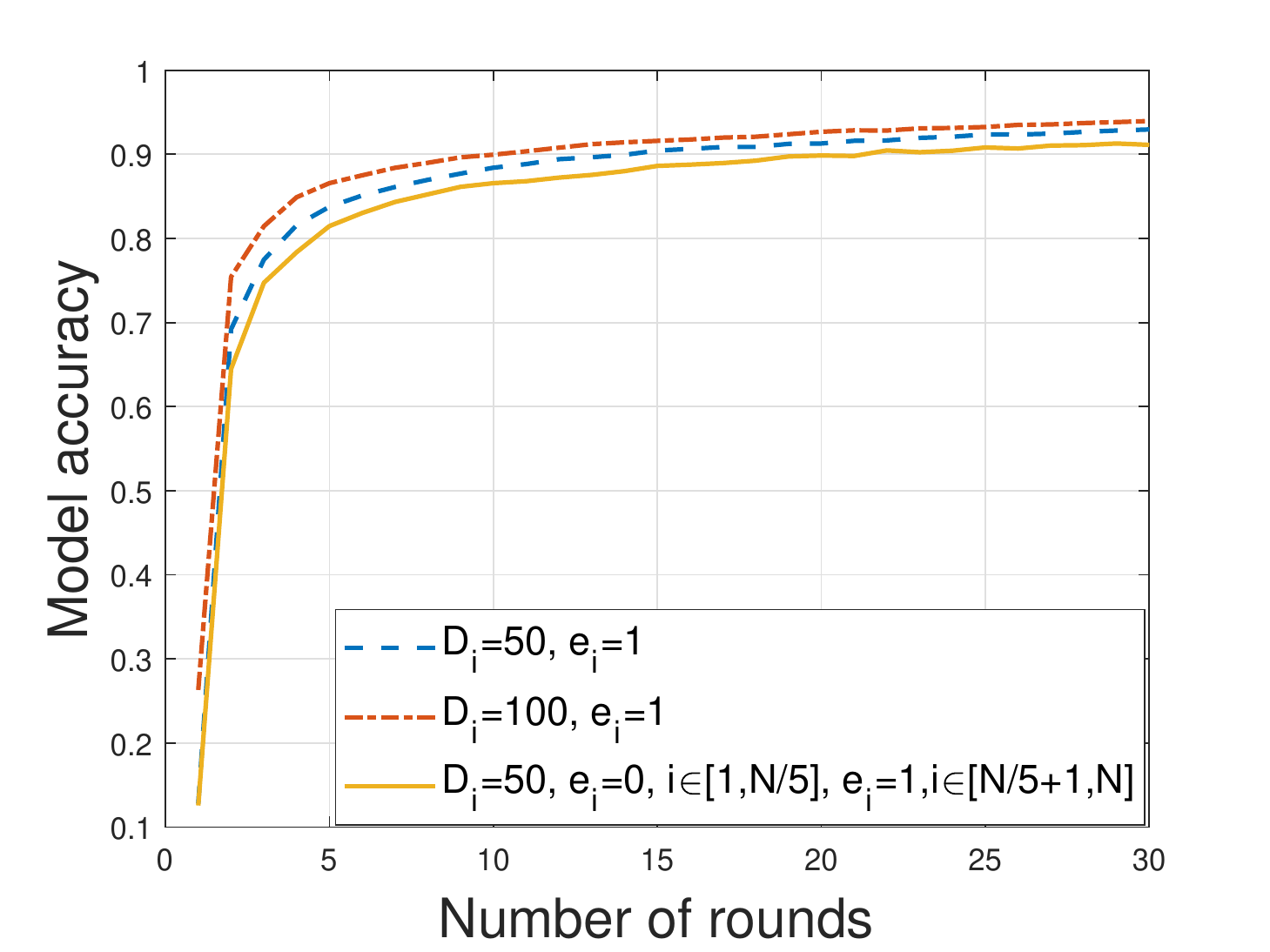}
\caption{Impact of effort level on the model accuracy.}
		\label{fig:ed_acc}
\end{minipage}
\hfill
\begin{minipage}[t]{.32\textwidth}
\centering
\includegraphics[width=1\textwidth]{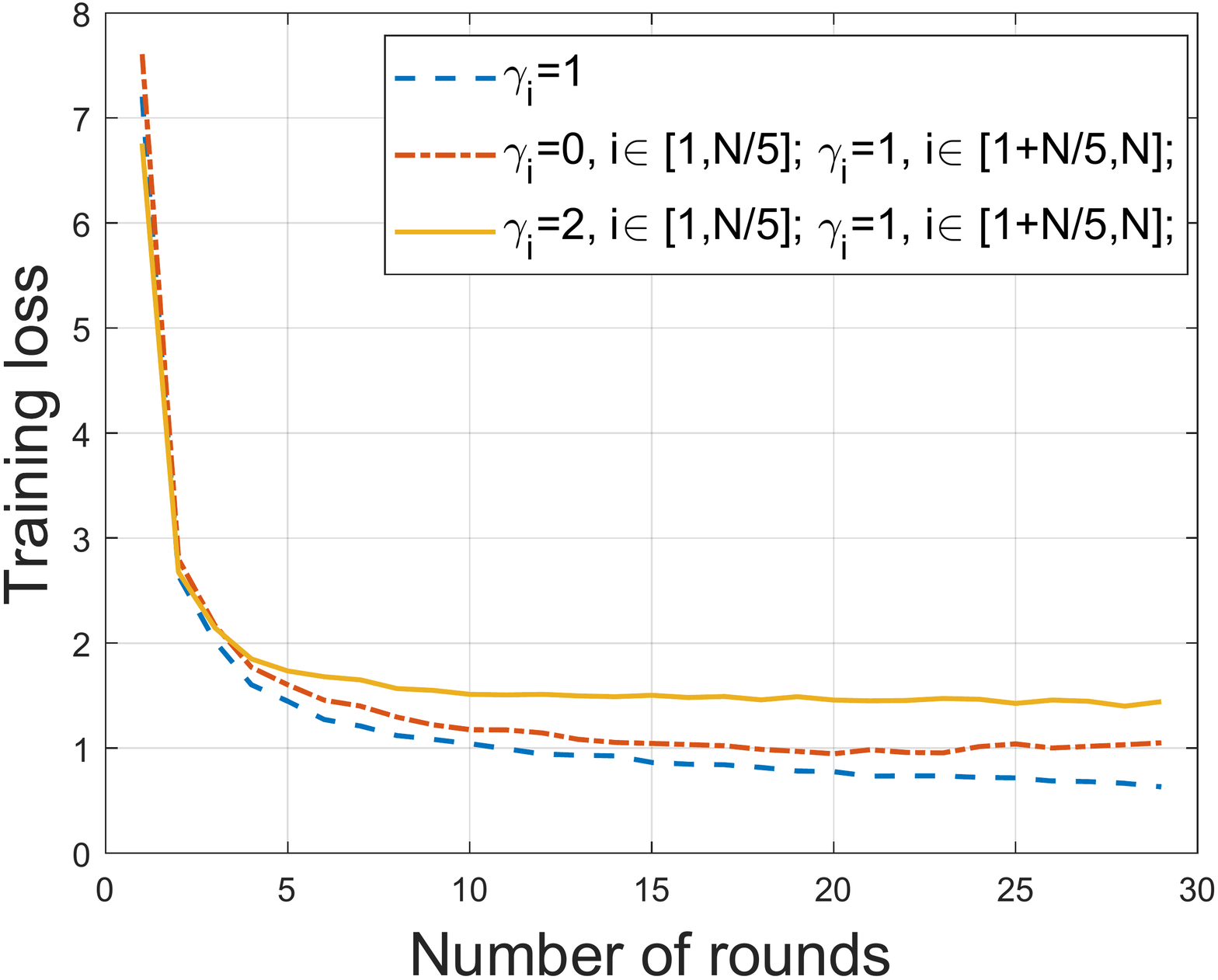}
\caption{Impact of model reporting coefficient on the training loss.}
		\label{fig:gamma_loss}
\end{minipage}
\vspace{-0.5cm}
\end{figure*}

\begin{figure*}[t]
\centering
\begin{minipage}[t]{.32\textwidth}
\centering
\includegraphics[width=1\textwidth]{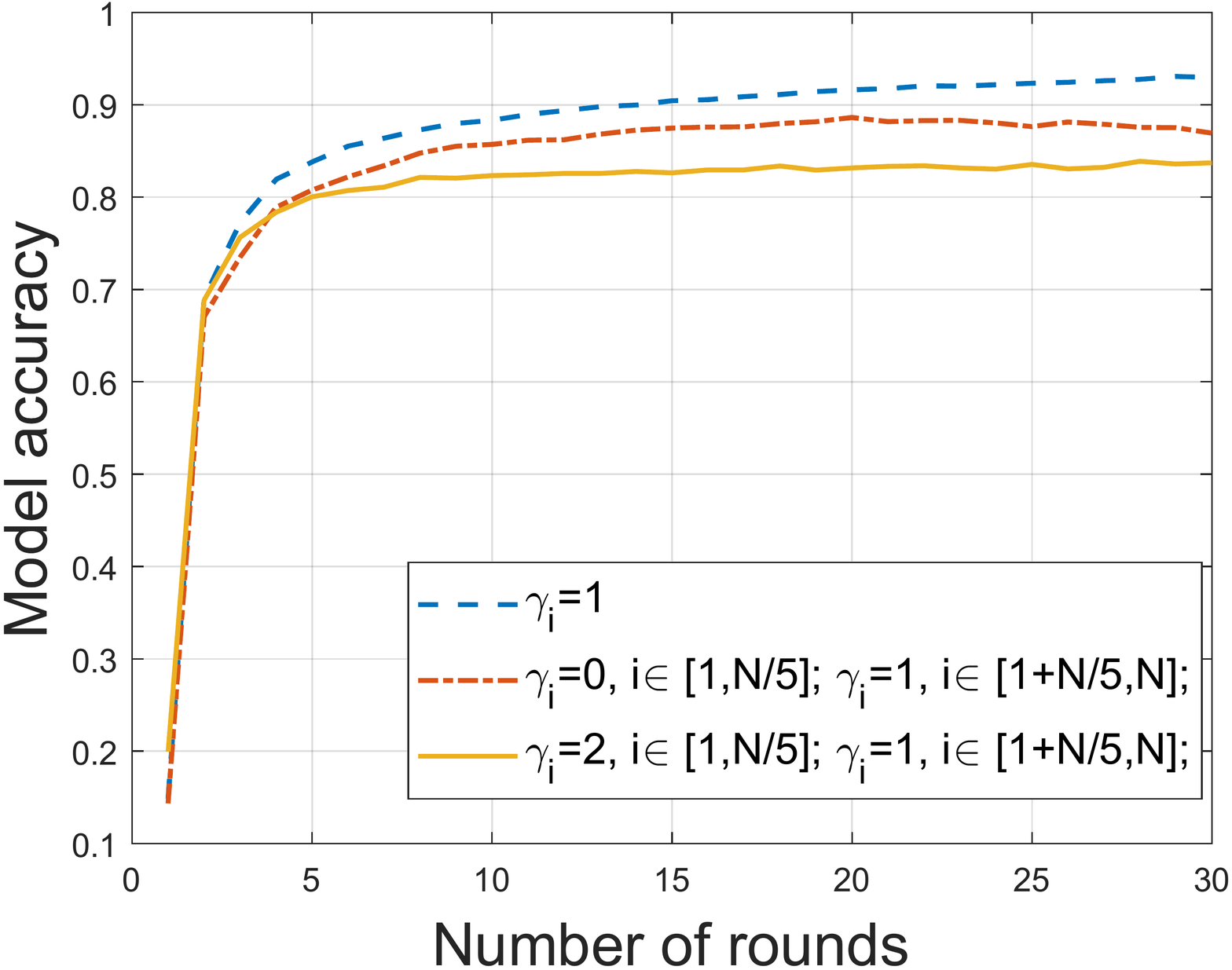}
\caption{Impact of model reporting coefficient on the model accuracy.}
		\label{fig:gamma_acc}
\end{minipage}
\hfill
\begin{minipage}[t]{.32\textwidth}
\centering
\includegraphics[width=1\textwidth]{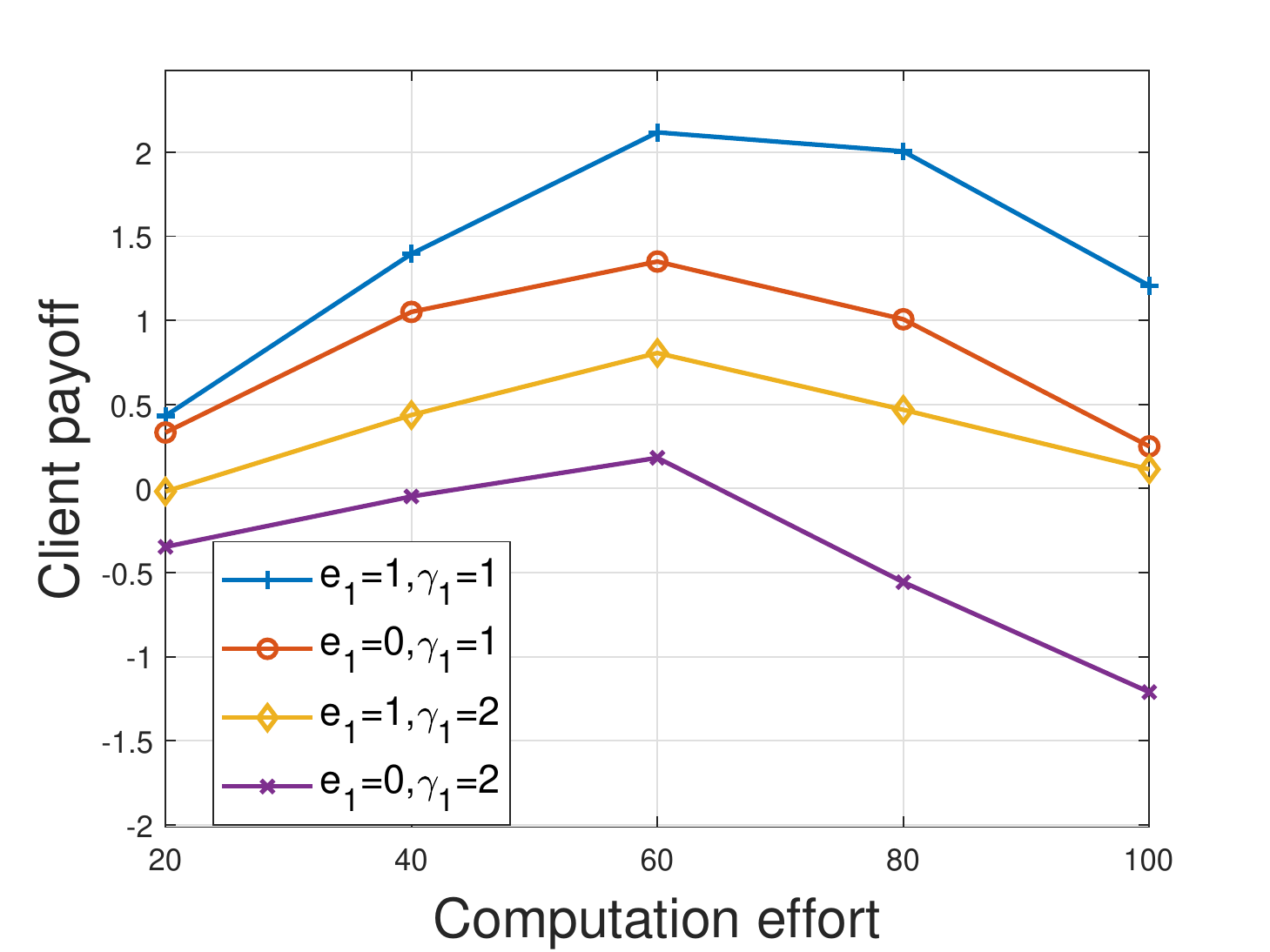}
\caption{Impact of clients' behavior on the payoff.}
		\label{fig:clientpay}
\end{minipage}
\hfill
\begin{minipage}[t]{.32\textwidth}
\centering
\includegraphics[width=1\textwidth]{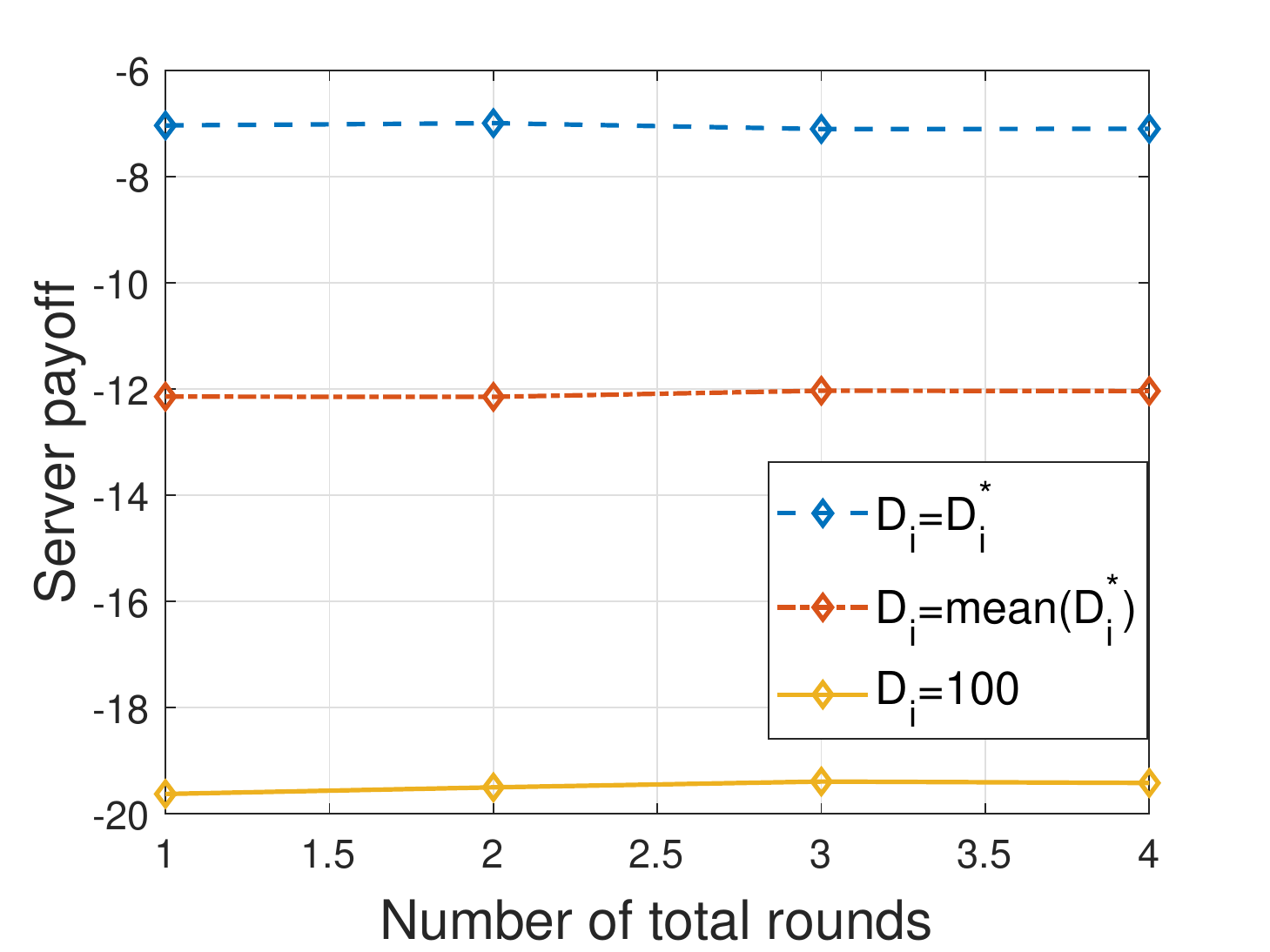}
\caption{Impact of computation effort allocation on server's payoff.}
		\label{fig:serverpay}
\end{minipage}
\vspace{-0.5cm}
\end{figure*}
In this section, we conduct real data based simulations to validate the theoretical findings and evaluate the LCEME mechanism. We first describe the simulation setups, and then we present the evaluation results and analyses.






 We implement a simulated system consisting of a server and 10 clients. We use the widely used MNIST dataset \cite{mnist} for simulations in Matlab. Each training element is a handwritten digit picture that represents numbers from 0 to 9. Each client conducts one layer of CNN for one local iteration in each round ($H=1$). We denote the heterogeneity degree of a client's dataset as the percentage of data with labels the same as the last digit of the client's index. For the remaining data of the client, we uniformly draw the training data samples from the entire training set. Unless otherwise specified, client $i$'s heterogeneity degree is $0.4$, and the mini-batch size is $D_i=50$.
\vspace{-0.2cm}
\subsection{Impact of Clients' Strategies on Training Loss}
We first compare the training loss while clients' data labeling and computation efforts changes. From Figs. \ref{fig:ed_loss} and \ref{fig:ed_acc}, we can see that the training loss decreases and the model accuracy increases as $D_i$ increases. We also observe that when there exist clients who make no effort in data labeling, the training loss increases, and the model accuracy decreases. The observations conform to our theoretical result in Theorem \ref{thm:loss}. We also compare the training loss while clients report local models with different model reporting coefficients and truthfully make efforts. We observe from Figs. \ref{fig:gamma_loss} and \ref{fig:gamma_acc} that the training loss is minimized when all clients report their actual local model. When there exist clients report local model untruthfully, the training loss increases, and the model accuracy decreases. This conforms to the result in Theorem \ref{thm:loss} that the more clients truthfully report local models, the lower the training loss. We also observe that, although the training loss bounds are the same when $\gamma_i=0$ and $\gamma_i=1$, the training loss is lower when $\gamma=0$. Figs. \ref{fig:ed_loss}, \ref{fig:ed_acc}, \ref{fig:gamma_loss}, and \ref{fig:gamma_acc} demonstrate that, when clients truthfully make efforts and report local models, the training loss is minimized and the model accuracy is maximized.
\vspace{-0.2cm}
\subsection{Impact of Truthfulness on Clients' Payoff}
We compare a client's payoff while making the desired data labeling effort $e_1=1$ or not $e_1=0$, and reporting the actual local model $\gamma_1=1$ or not $\gamma_1\neq 1$, as the computation effort $D_1$ changes. The assigned computation effort $D_1^\prime=60$. We let other clients behave truthfully. We observe from Fig. \ref{fig:clientpay} that a client's payoff, when she makes data labeling and computation effort as the server desired and reports actual local model, is always higher than that when her behavior is untruthful. Furthermore, we also observe that the client's payoff is positive when she behaves truthfully. The simulation results demonstrate that the LCEME mechanism is truthful and achieves the IR property.

	

\subsection{Server's Payoff}
We compare the server's payoff while clients make different computation efforts. From Fig. \ref{fig:serverpay}, we can see that the server's payoff is maximized when clients make the server's optimal computation effort. When clients do not make the optimal computation effort, the server's payoff is lower even if the total computation effort of clients is the same as the optimal computation effort allocation. This is because, in the former case, the computation effort allocation does not care about clients' heterogeneous computation cost and thus causes higher computation costs. We also simulate the case where clients' computation effort $D_i=100$ is always higher than the optimal computation effort. We observe that among three cases, this case results in the lowest server's payoff. This is because clients' computation costs are ignored when assigning $D_i$, resulting in an increase in the server cost.
\vspace{-0.2cm}


\section{Conclusion and Future Work}              \label{sc:concl}   \vspace{-0.2cm}

In this paper, we studied FL with crowdsourced data labels, where the local data of each participating client are labeled manually by the client. We characterized the performance bounds on the training loss as a function of clients' data labeling effort, local computation effort, and reported local models. We then devised truthful incentive mechanisms which motivate strategic clients to make truthful efforts as desired by the server in data labeling and local model computation, and also report true local models to the server based on the derived performance bound. Simulations based on real data demonstrated the efficacy of the proposed algorithms. 

{For future work, we will extend our study to more general settings. In this paper, we studied truthful mechanism design under the assumption that clients' costs are known to the server. The mechanism design problem where clients’ costs are also private is more practical but challenging. Another direction is to consider partial participation of clients. In this case, the truthful mechanism design and the optimal labeling effort assignment will be different.}
\vspace{-0.3cm}

%
%
\vspace{-0.2cm}
\section*{Appendix}
\vspace{-0.2cm}
\subsection{Proof of Theorem \ref{thm:loss}}\label{prf:loss}
\vspace{-0.2cm}
We define a virtual sequence ${\bar{\mathbf{w}}_{t,h}}$, given by
    $\bar{\mathbf{w}}_{t,h}= \sum_{i\in \mathcal{N}}p_i \mathbf{w}_{t,h}^{i}, \forall t,h.$
Note that ${\bar{\mathbf{w}}_{t,h}}$ is not accessible when clients have not completed $H$ local iterations (i.e., $h<H$), and $\mathbf{w}_{t}={\bar{\mathbf{w}}_{t,H}}$.
\vspace{-0.4cm}
\begin{equation}
\begin{aligned}
    &\left\|\bar{\mathbf{w}}_{t,H}-\mathbf{w}^{*}\right\|^{2}
    =\left\|\bar{\mathbf{w}}_{t,H-1}-\mathbf{w}^{*}-\eta \sum_{i\in \mathcal{N}}\gamma_ip_i{g_{t,H-1}^i}^\prime\right\|^{2}\leq
     \\&  2\underbrace{ \left\|\bar{\mathbf{w}}_{t,H-1}\!-\!\eta \bar{g}_{t,H-1}\!-\!\mathbf{w}^{*}\right\|^{2}}_{A_1}
    \!+2\underbrace{\left\|\eta \bar{g}_{t,H-1}\!-\!\eta \!\sum_{i\in \mathcal{N}}\gamma_ip_i{g_{t,H-1}^i}^\prime\right\|^{2}}_{A_2}
    \label{eq:A123}
\end{aligned}
\vspace{-0.1cm}
\end{equation}
where ${g_{t,h}^i}^\prime$ is the gradient when client $i$ makes any data labeling effort, $\bar{g}_{t,h}\triangleq\sum_{i\in \mathcal{N}}p_i\bar{g}^{i}_{t,h}\triangleq \sum_{i\in \mathcal{N}}p_iE[g_{t,h}^i]$, and ${g_{t,h}^i}$ is the gradient when client $i$ makes data labeling effort.
\vspace{-0.2cm}
  \begin{equation}
     \ba
    &A_{1}=
    \\&\left\|\bar{\mathbf{w}}_{t,H-1}-\mathbf{w}^{*}\right\|^{2}+\underbrace{\eta^{2} \|\bar{g}_{t,H-1}\|^{2}}_{B_1}-\underbrace{2 \eta\left\langle \bar{\mathbf{w}}_{t,H-1}-\mathbf{w}^{*}, \bar{g}_{t,H-1}\right\rangle}_{B_2}.
     \ea
     \label{eq:A1_noniid}
     \vspace{-0.2cm}
 \end{equation}
For $B_2$, we have
\vspace{-0.2cm}
\begin{displaymath}
    \ba
  B_2 
    &= -2\eta \sum_{i\in \mathcal{N}} p_i \langle \bar{\mathbf{w}}_{t,H-1}-{\mathbf{w}}_{t,H-1}^i,\bar{g}_{t,H-1}^i \rangle
    \\& -2\eta \sum_{i\in \mathcal{N}} p_i \langle {\mathbf{w}}_{t,H-1}^i-\mathbf{w}^*,\bar{g}_{t,H-1}^i \rangle.
    \ea
    \label{eq:B2_niid}
    \vspace{-0.2cm}
\end{displaymath}
We use the convexity of $\left\|\cdot\right\|^2$ and the $L$-smoothness of $F_i$ to bound $B_1$, the Cauchy-Schwarz inequality and AM-GM inequality to bound the first term of $B_2$, and the $\mu$-strong convexity of $F_i$ to bound the second term of $B_2$. We have
\vspace{-0.2cm}
\begin{align*}
      &A_{1}\leq \left\|\bar{\mathbf{w}}_{t,H-1}-\mathbf{w}^{*}\right\|^{2} + {2L\eta^2} \sum_{i\in \mathcal{N}} p_i(F_i(\mathbf{w}_{t,H-1}^i)-F_i(\mathbf{w}_i^*)) 
     \\&\ \ +{\sum_{i\in \mathcal{N}} p_i\left(\left\|\bar{\mathbf{w}}_{t,H-1}-{\mathbf{w}}_{t,H-1}^i\right\|^2+\eta^2\left\|\bar{g}_{t,H-1}^i\right\|^2\right)}
     \\&\ \ {-{2\eta}\sum_{i\in \mathcal{N}}p_i\left(F_{i}({\mathbf{w}}_{t,H-1}^{i})-F_{i}({\mathbf{w}}^{*})+\frac{\mu}{2}\left\|{\mathbf{w}}_{t,H-1}^{i}-{\mathbf{w}}^{*}\right\|^{2}\right)}
     \\& \leq (1-\mu\eta) \left\|\mathbf{w}_{t,H-1}-\mathbf{w}^{*}\right\|^{2}+\sum_{i\in \mathcal{N}}p_i\left\|\bar{\mathbf{w}}_{t,H-1}-{\mathbf{w}}_{t,H-1}^i\right\|^2
     \\& \ \  +{4L\eta^2\sum_{i\in \mathcal{N}} p_i(F_i(\mathbf{w}_{t,H-1}^i)-F_i(\mathbf{w}_i^*))}
     \\&\ \ {-2\eta \sum_{i\in \mathcal{N}} p_i \left(F_{i}({\mathbf{w}}_{t,H-1}^{i})-F_{i}({\mathbf{w}}^{*})\right)},
     \label{eq:A1_niid}
     \vspace{-0.2cm}
\end{align*}
 in which we denote the last two lines as $C_1$. 
 \vspace{-0.2cm}
  \begin{displaymath}
     \ba
       &C_1=4L\eta^2\sum_{i\in \mathcal{N}}p_i(F_i(\mathbf{w}^*)-F_i(\mathbf{w}_i^*))
      \\& - 2\eta (1-2L\eta)\sum_{i\in \mathcal{N}}p_i(F_i(\mathbf{w}_{t,H-1}^i)-F_i(\mathbf{w}^*))
      \\ \leq &  4L\eta^2\sum_{i\in \mathcal{N}}p_i d_i-2\eta(1-2L\eta)\left(-\sum_{i\in \mathcal{N}} p_i\right.
      \\& \left.(\eta L\left(F_i({\bar{\mathbf{w}}_{t,H-1}}-F_i(\mathbf{w}_i^*)\right)+\frac{1}{2\eta}\left\|{\mathbf{w}}_{t,H-1}^i-\bar{\mathbf{w}}_{t,H-1}\right\|^2\right.
    \\& \left.+F_i({\bar{\mathbf{w}}_{t,H-1}})-F_i(\mathbf{w}^*))
      \right). 
     \\ \leq & 2\eta(1-2L\eta)(\eta L-1)\sum_{i\in \mathcal{N}} p_i\left(F_i({\bar{\mathbf{w}}_{t,H-1}}-F_i(\mathbf{w}_i^*)\right)
     \\& +(4L\eta^2+2L\eta^2(1-2L\eta))\sum_{i\in \mathcal{N}}p_i d_i
     \\& +(1-2L\eta)\sum_{i\in \mathcal{N}}p_i\left\|{\mathbf{w}}_{t,H-1}^i-\bar{\mathbf{w}}_{t,H-1}\right\|^2
     \\ \leq &6L\eta^2\sum_{i\in \mathcal{N}}p_i d_i+\sum_{i\in \mathcal{N}}p_i\left\|{\mathbf{w}}_{t,H-1}^i-\bar{\mathbf{w}}_{t,H-1}\right\|^2.
     \ea
     \vspace{-0.2cm}
 \end{displaymath}
 Thus we can further bound $A_1$ as
\vspace{-0.2cm}
\begin{equation}
     \ba
     & E[A_1]
     \leq  (1-\mu\eta) \left\|\bar{\mathbf{w}}_{t,H-1}-\mathbf{w}^{*}\right\|^{2}
     \\& +6L\eta^2\sum_{i\in \mathcal{N}}p_i d_i+2\sum_{i\in \mathcal{N}}p_i\left\|{\mathbf{w}}_{t,H-1}^i-\bar{\mathbf{w}}_{t,H-1}\right\|^2.
     \ea
     \label{eq:eA1_niid}
     \vspace{-0.2cm}
 \end{equation}
 Next, we bound
 \vspace{-0.2cm}
  \begin{equation}
     \ba
      &\sum_{i\in \mathcal{N}}p_iE \left\|\bar{\mathbf{w}}_{t,h}-\mathbf{w}_{t,h}^{i}\right\|^{2}
      \leq  \sum_{i\in \mathcal{N}}p_iE \left\|\mathbf{w}_{t,h}^{i}-\mathbf{w}_{t,1}\right\|^{2}
      \\\leq& \eta^2\sum_{i\in \mathcal{N}}p_iE\|\sum_{h=1}^{H-1}{g_{t,h}^i}^\prime\|^2
       \leq  \eta^2(H-1)\sum_{i\in \mathcal{N}}p_i\sum_{h=1}^{H-1}E\|{g_{t,h}^i}^\prime\|^2.
     \ea
     \label{eq:model_var_miter_1}
     \vspace{-0.2cm}
 \end{equation}
 Using Assumption \ref{as:4}, we have
 \vspace{-0.2cm}
 \begin{align}
    &\ \ E\|g_{t,h}^i-{g_{t,h}^i}^\prime\|^2\nonumber
    \\&=E\|\frac{1}{D_i}\sum_{j}(\nabla f_i(\mathbf{w}_{t,h},\xi_{t}^{i,j})-\nabla f_i(\mathbf{w}_{t,h},{\xi_{t}^{i,j}}^\prime))\|^2\nonumber
    \\& \leq \frac{1}{D_i}\sum_{j}E_{{\xi_{t}^{i,j}}^\prime|\xi_{t}^{i,j}}\left[\|(\nabla f_i(\mathbf{w}_{t,h},\xi_{t}^{i,j})-\nabla f_i(\mathbf{w}_{t,h},{\xi_{t}^{i,j}}^\prime))\|^2\right]\nonumber
    \\& \leq{(1-e_i)}\beta. \label{eq:noe}
 \end{align}
 From \cite{dekel12optimal}, we have 
 \vspace{-0.2cm}
 \begin{equation}
     E\left\| \bar{g}_{t,h}^i-g_{t,h}^i\right\|^{2}\leq \frac{\sigma_i^2}{D_i}.
     \label{eq:A2_1_i}
     \vspace{-0.2cm}
 \end{equation}
From \eqref{eq:noe}, \eqref{eq:A2_1_i}, and Assumption 5, we have
\vspace{-0.2cm}
 \begin{align}
     &E\left\|{g_{t,h}^i}^\prime \right\|^2
     =E\left\|{g_{t,h}^i}^\prime -{g_{t,h}^i}+{g_{t,h}^i}-{\bar{g}_{t,h}^i}+{\bar{g}_{t,h}^i}\right\|^2\nonumber
     \\\leq &2E\left\|{g_{t,h}^i}^\prime -{g_{t,h}^i}\right\|^2+2E\left\|{g_{t,h}^i}-{\bar{g}_{t,h}^i}\right\|^2+2E\left\|{\bar{g}_{t,h}^i}\right\|^2\nonumber
     \\ \leq & 2(1-e_i)\beta+\frac{2\sigma_i^2}{D_i}+2G^2. \label{eq:A2_2}
 \end{align}

 Thus we can bound \eqref{eq:model_var_miter_1} as
 \vspace{-0.2cm}
 \begin{equation}
     \ba
      &\sum_{i\in \mathcal{N}}p_iE \left\|\bar{\mathbf{w}}_{t,h}-\mathbf{w}_{t,h}^{i}\right\|^{2}
      \\ \leq & 2\eta^2(H-1)^2\sum_{i\in \mathcal{N}}p_i((1-e_i)\beta+\frac{\sigma_i^2}{D_i}+G^2).
     \ea
     \label{eq:model_var_miter}
     \vspace{-0.2cm}
 \end{equation}
 Next, we bound $A_2$. From \eqref{eq:A2_1_i} and \eqref{eq:A2_2}, we have
 \vspace{-0.2cm}
  \begin{align}
    
     &E[A_2]
     =\left\|\eta \bar{g}_{t,H-1}-\eta \sum_{i\in \mathcal{N}}\gamma_ip_i{g_{t,H-1}^i}^\prime\right\|^{2}\nonumber
     \\= & \eta^2E\left\| \bar{g}_{t,H-1}-g_{t,H-1}+g_{t,H-1}+\sum_{i\in \mathcal{N}}\gamma_ip_i{g_{t,H-1}^i}^\prime\right\|^{2}\nonumber
     \\\leq & 2\eta^2E\left\| \bar{g}_{t,H-1}-g_{t,H-1}\right\|^{2}+2\eta^2E\left\|g_{t,H-1}^\prime -g_{t,H-1}\right\|^{2}\nonumber
     \\&+2\eta^2\sum_{i\in \mathcal{N}}p_i (\gamma_i-1)^2E\left\|{g_{t,H-1}^i}^\prime \right\|^2\nonumber
       \\\leq& 2{\eta^2} \sum_{i\in \mathcal{N}}({p_i}^2\frac{\sigma_i^2}{D_i}+{p_i}{(1-e_i)}\beta\nonumber
     \\ &+2p_i(\gamma_i-1)^2(G^2+\frac{\sigma_i^2}{D_i}+{(1-e_i)}\beta)) 
     .
     \label{eq:A2_niid}
     \vspace{-0.2cm}
 \end{align}
 
 
Combining (\ref{eq:A123}), (\ref{eq:eA1_niid}), \eqref{eq:model_var_miter}, and (\ref{eq:A2_niid}), we have
\vspace{-0.2cm}
  \begin{displaymath}
     \ba
        & E\left\|\mathbf{w}_{T,H}-  \mathbf{w}^{*}\right\|^{2} 
      \\ \leq &2(1-\mu\eta) \left\|{\mathbf{w}}_{T,H-1}-\mathbf{w}^{*}\right\|^{2} +12L\eta^2\sum_{i\in \mathcal{N}} p_i  d_i
      \\&+4{\eta^2} \sum_{i\in \mathcal{N}}({p_i}^2\frac{\sigma_i^2}{D_i}+{p_i}{(1-e_i)}\beta)
      \\& +4\eta^2\sum_{i\in \mathcal{N}}p_i((\gamma_i-1)^2+2(H-1)^2)(G^2+\frac{\sigma_i^2}{D_i}+{(1-e_i)}\beta).
     \ea
     \vspace{-0.2cm}
\end{displaymath}

Using induction and the smoothness of $F$, we have \eqref{eq:loss}.
\vspace{-0.2cm}
\subsection{Proof of Theorem \ref{thm:IR}}\label{prf:IR}
 Given that all users behave truthfully, the expected payoff of user $i$, $\forall i$ is given by
 \vspace{-0.2cm}
\begin{displaymath}
\ba
&E[u_i]
= \Omega(\boldsymbol{D}^\prime)-\Phi(D_i^\prime)F(\mathbf{w}_T)+c_l-c_le_i^\prime-Tc_p^iD_i^\prime.
   \\ \geq & \Phi(D_i^\prime) (F(\mathbf{w}_T)-F(\mathbf{w}^*))+Tc_p^iD_i^\prime
   \\&-\Phi(D_i^\prime)(F(\mathbf{w}_T)-F(\mathbf{w}^*))+c_l-c_le_i^\prime-Tc_p^iD_i^\prime
    =  0.
\ea
\vspace{-0.2cm}
\end{displaymath}

\vspace{-0.2cm}

\subsection{Proof of Theorem \ref{thm:SO}}\label{prf:so}
The total expected reward paid by the server is bounded by $\sum_{i\in\mathcal{N}}r_i \geq \sum_{i\in \mathcal{N} } (c_l+Tc_p^iD_i).$ Using \eqref{eq:loss}, we have
\vspace{-0.2cm}
\begin{displaymath}
\ba
    & F(\mathbf{w}_{T})-F(\mathbf{w}^*)+\sum_{i\in \mathcal{N} } r_i
     \leq {L}(1-\mu \eta)^{TH} E\left\|\mathbf{w}_{0}-\mathbf{w}^{*}\right\|^{2}
    \\&+
      {\sum_{i\in {\mathcal{N}}}(A({p_i}^2\frac{\sigma_i^2}{D_{t}^i}}+6L p_id_i+2p_i(H-1)^2\frac{\sigma_i^2}{D_{t}^i})+c_l+Tc_p^iD_i).
\ea
\vspace{-0.2cm}
\end{displaymath}

It can be shown that the above upper bound is a convex function of $D_i$. The optimal mini-batch size ${D_i}^*$ can be obtained by calculating the partial derivative of the bound with respect to $D_i$ and letting the derivative equals to 0.




\ifCLASSOPTIONcaptionsoff
  \newpage
\fi

\end{document}